\documentclass[conference]{IEEEtran}
\IEEEoverridecommandlockouts

\usepackage{cite}
\usepackage{amsmath,amssymb,amsfonts}
\usepackage{algorithmic}
\usepackage{graphicx}
\usepackage{textcomp}
\usepackage{xcolor}
\usepackage{caption}
\usepackage{subcaption}
\usepackage{array,multirow}
\def\BibTeX{{\rm B\kern-.05em{\sc i\kern-.025em b}\kern-.08em
    T\kern-.1667em\lower.7ex\hbox{E}\kern-.125emX}}
\begin{document}

\title{A Study of Data Augmentation Techniques to Overcome Data Scarcity in Wound Classification using Deep Learning}

\author{\IEEEauthorblockN{Harini Narayanan}
\IEEEauthorblockA{
\textit{BASIS Independent Fremont}\\
Fremont, CA \\
harini.sowmi.bb@gmail.com}
\and
\IEEEauthorblockN{Sindhu Ghanta}
\IEEEauthorblockA{AIClub Research Institute\\
San Jose, CA \\
sindhu@aiclub.world}}

\maketitle

\begin{abstract}

Chronic wounds are a significant burden on individuals and the healthcare system, affecting millions of people and incurring high costs. Wound classification using deep learning techniques is a promising approach for faster diagnosis and treatment initiation. However, lack of high quality data to train the ML models is a major challenge to realize the potential of ML in wound care. In fact, data limitations are the biggest challenge in studies using medical or forensic imaging today. We study data augmentation techniques that can be used to overcome the data scarcity limitations and unlock the potential of deep learning based solutions. In our study we explore a range of data augmentation techniques from geometric transformations of wound images to advanced GANs, to enrich and expand datasets. Using the Keras, Tensorflow, and Pandas libraries, we implemented the data augmentation techniques that can generate realistic wound images. We show that geometric data augmentation can improve classification performance, F1 scores, by up to 11\% on top of state-of-the-art models, across several key classes of wounds.  Our experiments with GAN based augmentation prove the viability of using DE-GANs to generate wound images with richer variations.  Our study and results show that data augmentation is a valuable privacy-preserving tool with huge potential to overcome the data scarcity limitations and we believe it will be part of any real-world ML-based wound care system.
 
\end{abstract}


\section{Introduction}
Wounds, the modern day focus of extensive medical and forensic research,
play an important role in quality of life, causing an immense amount of complications, ranging from small to large-scale. A chronic wound is defined as a wound that has been open for more than a month and that has not healed normally \cite{sen2021human}. Chronic wounds are a significant burden on individuals and the healthcare system, affecting millions of people and incurring high costs. Most common in elderly people, they affect around 8.2 million medicare beneficiaries in the United States, which shows the astounding ubiquity of wounds \cite{sen2021human}. 

Chronic wound classes include, but are not limited to, diabetic, pressure, surgical, and venous ulcers. Each of these types of wounds are created by a multitude of factors, such as diabetes, surgery, and venous insufficiencies. More than being painful and inconvenient, chronic wounds are a financial burden as well. Costs of individual patient care alone for pressure ulcers is \$20,900 minimum for each patient, with additional costs amounting to \$43,180 per year \cite{sen2021human}. Given the substantial impact a single wound can have on an individual's quality of life, the cumulative effect on the larger population diagnosed with chronic wounds is significant. Therefore, developing methods and systems for efficient, effective and accurate wound care are of tremendous importance to society.

The length and complexity of the medical care process often lead to individuals avoiding medical consultation, exacerbating their wound's condition. Automated techniques based on deep learning algorithms can be used to classify wounds, allowing for quicker diagnosis and treatment initiation. They can assist doctors in saving time and focusing on more critical steps of wound care. ML(machine learning)-based wound classification can improve the quality of life in many people around the world.

Many studies have concluded that application of automated techniques in wound classification and care is essential for the forensic and medical industry's improvements \cite{chan2022clinical}, \cite{dabas2023application}, \cite{reifs2023clinical}. However, large amounts of labeled data is required for developing deep learning models that can help solve wound classification. Collecting large amounts of real patient wound images is often
not feasible due to a variety of privacy and legal concerns. In almost
all use of deep learning techniques in forensic and medical sciences, there is the limitation
of not having enough readily available images or data to train models -- this is also called out in the conclusions of these wound classificiation studies  \cite{sarp2021enlightening}, \cite{rostami2021multiclass}, and \cite{anisuzzaman2022multi}). Such data limitations are a major challenge in realizing the potential that these ML techniques have in transforming wound care. 

Our study investigates the potential of state-of-the-art data augmentation techniques to overcome the data scarcity limitations in building real-world wound care systems. We explore two categories of data augmentation techniques, one based on geometric transformations and the other based on generative adversarial networks (GAN) \cite{goodfellow2014generative}. Both can generate realistic wound images which can then be used to train deep learning models that achieve higher wound classification accuracy.  Our study uses state-of-the-art computer vision (CV) models and transfer learning\cite{wiki:Transfer_learning} as a baseline to mimic real-world ML-based wound care applications -- as a part of this, we also show how modern CV models can be adapted to wound classification via transfer learning.  Our results show that geometric data augmentation can provide significant improvements (up to 11\% in F1 scores) on top of state-of-the-art models, and prove the viability of using GAN-based models to generate richer wound images. 

In the next section we describe various research studies that have shown the viability of deep learning for wound care, the applicability of data augmentation, and the broad limitations arising from lack of good labeled data. In Section \ref{sec:methods} we outline the datasets we use, the augmentation and transfer learning\cite{wiki:Transfer_learning} techniques we have implemented and evaluated. Section \ref{sec:experiments} summarizes our key results and we discuss these results in Section \ref{sec:discussion}. The final section provides a conclusion and outlines future work.

\section{Background}
\label{sec:background}
Deep learning classification algorithms have shown immense success in medical imaging in recent years, including in the classification of chronic wounds \cite{chan2022clinical}, \cite{dabas2023application}, \cite{reifs2023clinical}.  Studies including  \cite{chan2022clinical}, \cite{dabas2023application}, and \cite{reifs2023clinical} have concluded that the application of deep learning algorithms to wound assessment and patient care is viable and essential to the growing industry. These studies have also asserted that automated techniques for wound classification will lead to faster diagnosis and treatment. This opened up an area vast in opportunities to further research on chronic wound classification and diagnosis. 

After these viability proofs, there were many studies that demonstrated
basic wound classification with different ML models. Rostami et. al \cite{rostami2021multiclass} used a Convolutional Neural Network (CNN) that classified between surgical, diabetic, and venous ulcers with high accuracies. 
Similar research conducted by Sarp et al. \cite{sarp2021enlightening}, used transfer
learning and XAI models that classified chronic wounds, similar to the
research performed by Anisuzzaman et al. \cite{anisuzzaman2022wound}. Chairat et al. \cite{chairat2023ai}
worked on increasing the accessibility of wound care by using transfer
learning models such as U-Net with EfficientNet and U-Net with
MobileNetV2 \cite{movilenetv2} to assess a wound and determine the rate of wound healing and 
made wound care more accessible to the general public by
training their models on images taken by an iPhone at different
locations, times of day, and brightness. Taking this one step further, subsequent studies went beyond just
classifying wounds and looked at wound analysis such as wound severity
and healing rate \cite{anisuzzaman2022multi, husers2022automatic, gupta2023towards}.

All these studies demonstrate the huge promise of deep learning models to transform wound care. However, data scarcity is a huge limitation in practical use of such techniques in wound care. This is explicitly stated in the conclusions of studies \cite{sarp2021enlightening}, \cite{rostami2021multiclass}, and \cite{anisuzzaman2022multi}, and in the discussions written by studies \cite{anisuzzaman2022multi} and \cite{husers2022automatic}. For example, both \cite{anisuzzaman2022multi} and \cite{husers2022automatic} described in their future work that their model
should be validated on a larger dataset. An efficient and scalable way
to solve this problem is essential for continued research and successful
application of deep learning techniques in the field of wound care. Data augmentation provides a scalable and privacy preserving technique to overcome this data scarcity. 

Broadly, data augmentation has been shown to improve deep learning models \cite{shorten2019survey}. For wound care, it has been explored in the context of wound classification, segmentation, and infection prediction. Liu et al. \cite{Liu2024} demonstrate the use of semi-supervised learning with a secondary dataset to augment a wound dataset to improve wound classification. 
Several researchers have explored the use of data augmentation to improve wound segmentation \cite{Gutbrad2024, Niri2021, arturas2023}. Gutbrad et al. \cite{Gutbrad2024} explore warping of wound edges and blurring of shapes to augment foot wound images to improve wound segmentation, Niri et al. 
\cite{Niri2021} have shown the effectiveness of a multi-view data augmentation scheme in improving wound segmentation, and  Kairys and Vidas \cite{arturas2023} explore the use of photometric and geometric augmentation techniques for diabetic foot ulcer semantic segmentation. Zhao et al. \cite{Zhao2023} use a multivariate time series based data augmentation method to predict the infection of wounds. Researchers have also used general augmentation strategies such as CutMmix \cite{yun2019cutmixregularizationstrategytrain} to augment data sets for building explainable AI models for vascular wound images \cite{Zhiwen2024}. 

Our study takes a pragmatic view point and evaluates data augmentation techniques that can be used to build real-world deep learning based wound care systems. Such real-world systems do not start from scratch and tend to build on top of  state-of-the-art  computer vision (CV) models -- to mimic this we have selected three widely used CV models as foundation models, viz., MobileNet V2\cite{movilenetv2}, ResNet50\cite{Resnet50}, and VGG16\cite{vgg16}. We apply transfer learning\cite{wiki:Transfer_learning} to adapt these models for wound classification.  We then use these as baseline models and show that using geometric data augmentation we can achieve significant classification performance improvement (up to 11\% in F1 scores) on top of this baseline across key wound classes. We also demonstrate the viability of using GANs to generate wound images with richer variations.

\section{Methods}
\label{sec:methods}

\begin{figure}
    \centering
    \includegraphics[scale = 0.25]{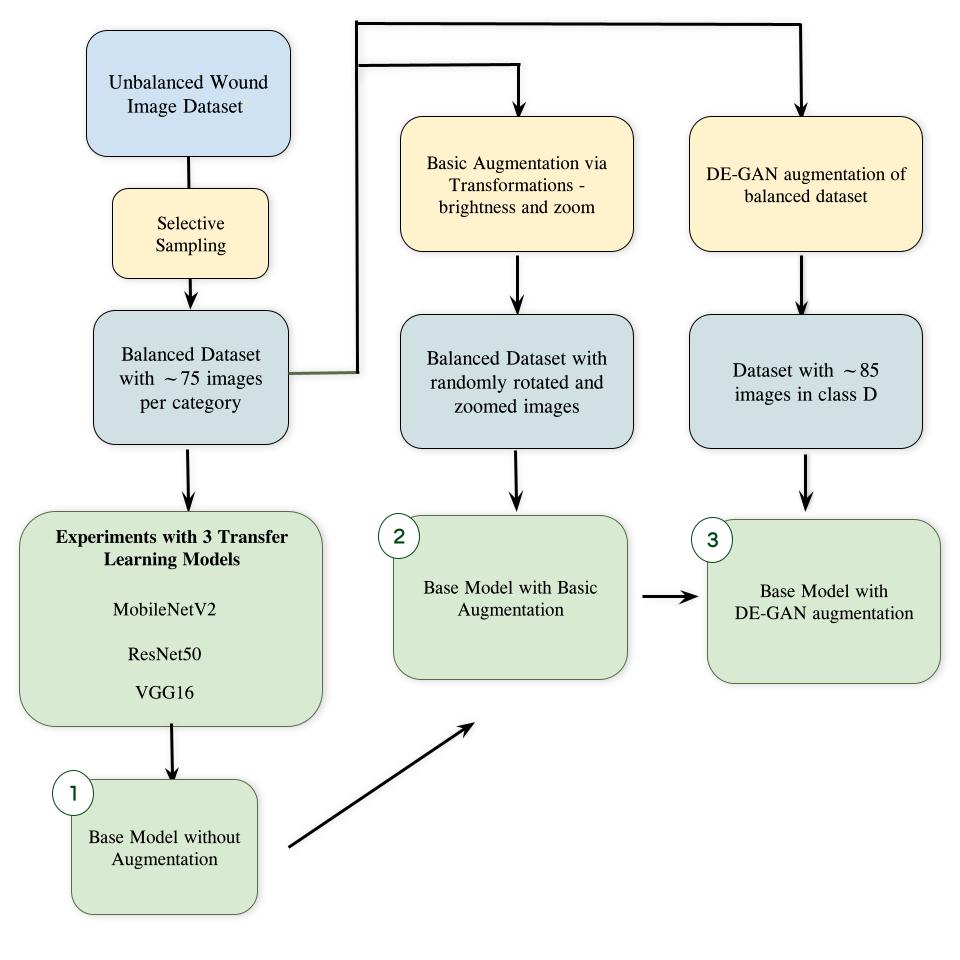}
    \caption{A flow chart representing the overall methodology employed in the study}
    \label{fig:overall methods}
\end{figure}

This section describes the three major phases in our experiments: transfer learning\cite{wiki:Transfer_learning} from foundational CV models, geometric data augmentation, and GAN based data augmentation. Figure \ref{fig:overall methods} visually shows the key steps and the three phases. We begin by describing the composition of the dataset, as well as how the dataset is split. This is followed by an outline of the training process, and detailed descriptions of the base model and data augmentation techniques implemented. 

\subsection{Dataset}

\begin{table}[htbp]
\caption{Original Dataset with Number of Data Points}
\begin{center}
\begin{tabular}{|c|c|c|c|}
\hline
\textbf{Wound Type} & \textbf{Number of images} \\
\hline
Venous (class V) & 185 images \\
\hline
Background (class BG) & 75 images \\
\hline
Diabetic (class D) & 139 images \\
\hline
Not an Ulcer (class N) & 75 images \\
\hline
Pressure (class P) & 100 images \\
\hline
Surgical (class S) & 122 images \\
\hline
\end{tabular}
\end{center}
\end{table}

\begin{figure}
    \centering
    \includegraphics[scale = 0.27]{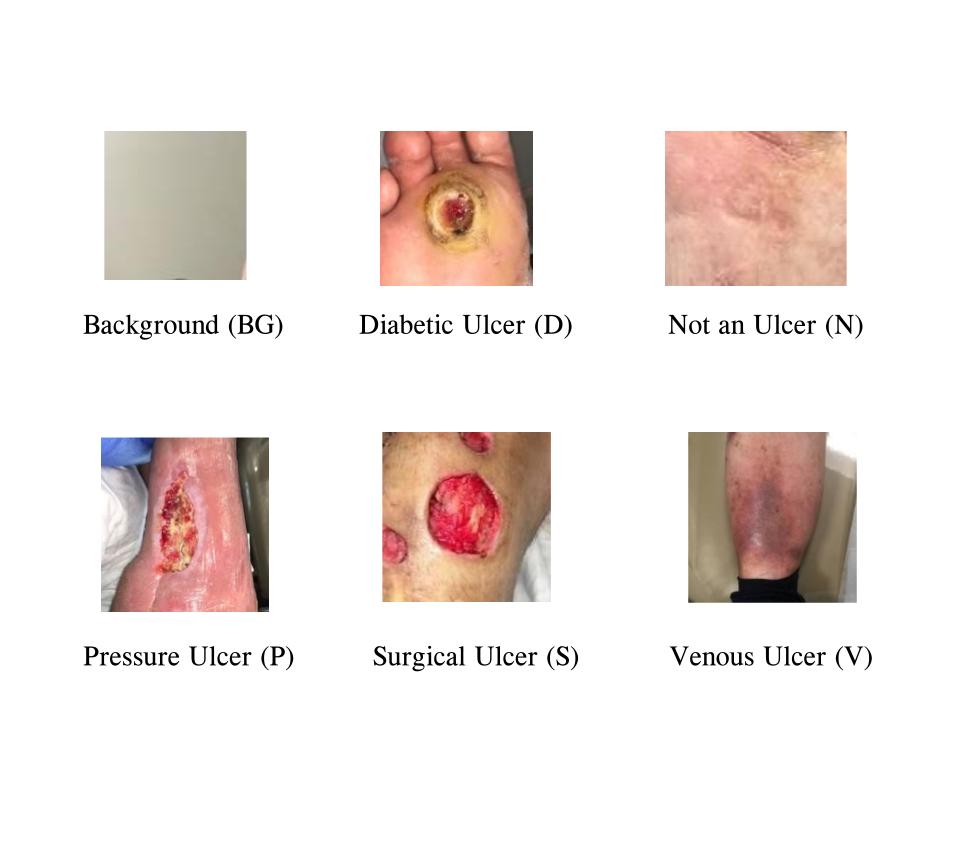}
    \caption{Examples of all categories of wounds addressed in this study}
    \label{fig:1}
\end{figure}

\begin{figure}
    \centering
    \includegraphics[scale = 0.3]{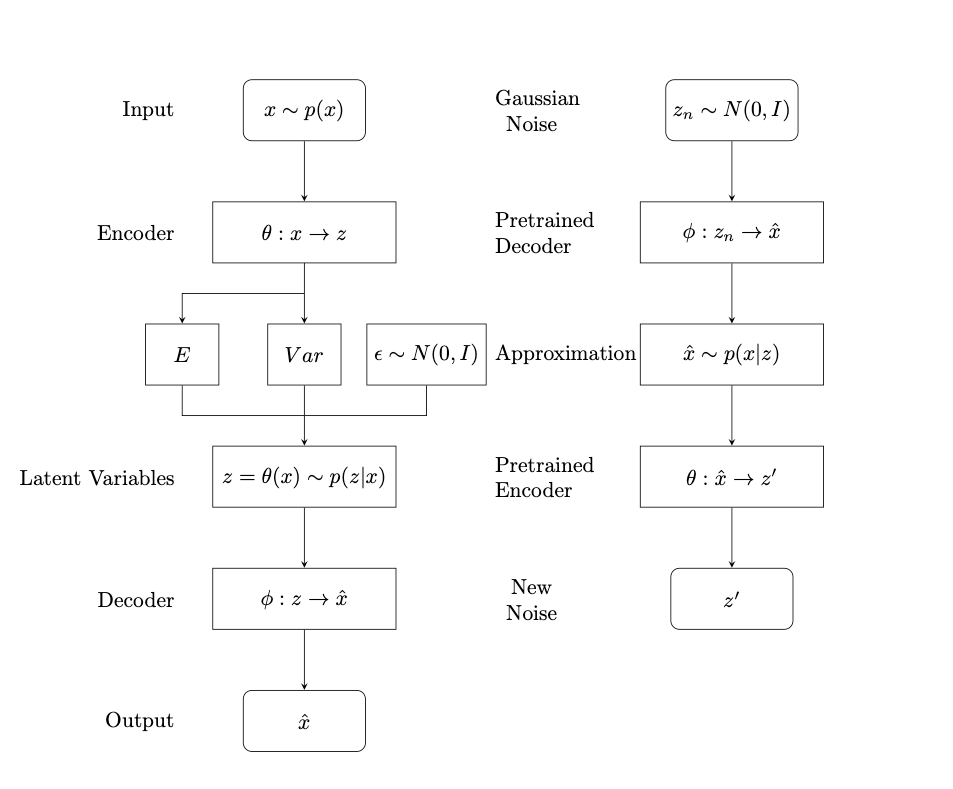}
    \caption{A Flow Chart Representing the Parts of a DE-GAN \cite{zhong2020generative}
\emph{Note. }The right diagram represents the structure of the
variational autoencoder in the DE-GAN, and the left diagram represents
the structure of the naive GAN component of the DE-GAN.}
    \label{fig:structure of gan}
\end{figure}

The original dataset, obtained from another study \cite{anisuzzaman2022wound}, had a very imbalanced dataset --some classes having as many as 2 times more images than others, as shown in Table 1. These classes included, background (BG), diabetic ulcers (D), not an ulcer (N), pressure ulcers (P), surgical ulcers (S), and venous ulcers (V). Initially, the category with the most images was the venous ulcers class, and the not an ulcer class had the least amount of images. Selective sampling was used to create a balanced dataset. This was accomplished by randomly choosing about 75 images from each category. Each one of these wound types have a very distinct appearance, as shown in Figure \ref{fig:1}. We split our data into separate train and test sets with an 80-20 split, respectively. 

\subsection{Wound classification models using transfer learning}
Over the years, a class of foundational computer vision models have
emerged. These foundational models such as ResNet50\cite{Resnet50}, VGG16\cite{vgg16}, MobileNetV2\cite{movilenetv2},
can be adapted to domain specific tasks such as wound classification, by
applying transfer learning. This study tests three of these foundational models with a vast range of hyperparameters to select the one that is most accurate (see Figure \ref{fig:overall methods}). Different combinations between 10, 30, and 50 epochs and learning rates of 0.01, 0.001, 0.0001, 0.05, and 0.0005 were also used. Though these models have high accuracy and are highly adaptable to specific tasks, due to the small dataset size, even the best models prove to have overfitting on the dataset.

\subsection{Geometric Data Augmentation}
An effective method to combat
overfitting is data augmentation. In this study, geometric data augmentation was implemented by using multiple python libraries. Tensorflow was used to implement transformations on the train dataset that rotate images by a degree between 0 and 30, and change the brightness of an image to a value ranging from 0 to 0.2 randomly. The balanced dataset underwent these transformations to yield a new dataset with the same number of images, but with each image rotated or dimmed/brightened randomly. The most accurate model from phase 1 of experiments was then trained with this new, transformed dataset. These variations force the model to learn the features of the dataset, which helps combat overfitting. Although geometric data augmentation is effective in increasing the accuracy of the model to some extent, it is limited in variations it can produce. Generative modeling, such as GAN, can be used to overcome this limitation to create new, varied images for the dataset.

\subsection{GAN-based data augmentation}
\label{subsec:gan-aug}
We use Generative Adversarial Networks (GAN) with Decoder-Encoder Output Noise (DE-GAN) introduced by \cite{zhong2020generative} with the built-in loss function. We trained the DE-GAN on the training data set to create synthetic diabetic ulcer images to expand the training dataset (see Figure \ref{fig:structure of gan} and Figure \ref{fig:loss function}). These were used to introduce variation and combat overfitting and confusion between classes, and they have also been proven to create higher quality images in less time than the naive GAN. Many hyperparameter combinations were tested, from 3000-6000 epochs and many learning rates ranging from 0.0005 to 0.1. The images generated by the DE-GANs were then used to inflate the original training dataset for class D by 14 images. This new dataset was used to train the most accurate foundation model from our initial experiments. 

\begin{figure}
    \centering
    {\Large
    \[ \mathcal{L} = \lambda_1 \mathcal{L}_{adv} + \lambda_2 \mathcal{L}_{hid} \]
    }
    \caption{The Combined Loss Function Used by the DE-GAN as described in
\cite{zhong2020generative}}
    \label{fig:loss function}
\end{figure}

\section{Experiments and Results}
\label{sec:experiments}
The goals of our experiments are
\begin{itemize}
\item Show transfer learning can be applied to adapt foundation CV models to wound classification
\item Show that geometric data augmentation techniques improve accuracy
\item Investigate the viability of DE-GAN techniques to successfully augment wound datasets and improve accuracy. 
\end{itemize}

The results from the experiments run using the two types of data augmentation are reported and analyzed. A set of models were trained using the balanced dataset with 75 images per category and transfer learning from the base models. Three different sizes (small, medium, large) of base models (MobileNetV2, ResNet50, and VGG16, respectively) were chosen in order to provide for a good coverage of the sizes of models used. After experimenting with more than 15 different combinations of learning rates and number of epochs, a few top scores listed in table 2 emerged. The two models with top classification accuracies were VGG16 and ResNet50, with about 82\% and 84\% accuracies, respectively.

\subsection{Results with data augmentation}
\begin{figure}
    \centering
    \includegraphics[scale= 0.25]{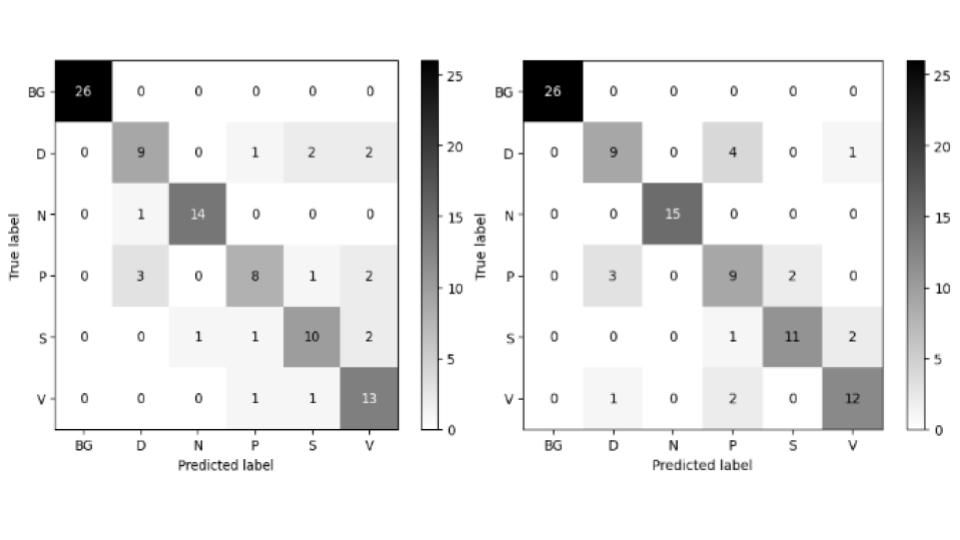}
    \caption{Confusion Matrix of Top Two Models -
    \emph{Note.} The right image shows the confusion matrix of the base
model VGG16 with 30 epochs and 0.0005 learning rate. The left image
shows the confusion matrix of the base model ResNet50 with 37 epochs and
0.0005 learning rate.}
    \label{fig:top two initial}
\end{figure}

Although these models performed quite well with transfer learning on the
original dataset with balancing, their confusion matrices shown in
figure \ref{fig:top two initial} show that they have very low accuracy for class D and class P. This demonstrates one of the key limitations of applying transfer learning, which is that foundation models still need a good amount and variety of data to adapt to a new domain such as wound classification. To address this core problem, we explored data augmentation to combat overfitting and increase accuracy.  

Geometric data augmentation provides significant improvement for classes D, P and V as shown in columns \textbf{Resnet50 with xfer learning} and \textbf{Geometric aug} in Table \ref{tab:precison recall and f1 scores}.  We see that geometric augmentation provides \textbf{10\%, 11\%}, and \textbf{7\%}  improvements in \textbf{F1} scores for classes D,  P, and V, respectively (see Table III) and these are precisely the classes that had low F1 scores with just transfer learning. This shows the power of Geometric data augmentation to improve upon transfer learning on state-of-the art CV models. 

\begin{table}[htbp]
\caption{Top accuracies after transfer learning and without data augmentation}
\begin{center}
\begin{tabular}{|c|c|c|c|}
\hline
\textbf{Model} & \textbf{Epochs} & \textbf{Learning Rate} &
\textbf{Accuracy} \\
\hline
MobileNetV2 & 30 & 0.0005 & 0.64 \\
\hline
VGG16 & 30 & 0.001 & 0.82 \\
\hline
\textbf{ResNet50} & \textbf{37} & \textbf{0.0005} &
\textbf{0.84} \\
\hline
\end{tabular}
\end{center}
\end{table}

\begin{table}[htbp]
\caption{Precision, Recall and F1-scores for Foundation model (ResNet50) with transfer learning, geometric augmentation and DE-GAN augmentation. Note the significant F1-score improvement for classes D, P, and V.}
\label{tab:precison recall and f1 scores}
\begin{center}
\begin{tabular}{|c|l|c|c|c|} \hline  
\textbf{Wound}& \textbf{Scores} & \textbf{ResNet50 with}& \textbf{Geometric}& \textbf{DE-GAN}\\
 \textbf{Classes}& & \textbf{xfer learning}& \textbf{aug}&\textbf{aug}\\ \hline  
 &Precision& 1.0& 1.0&1.0\\ 
  BG&Recall& 1.0& 0.93&0.93\\ 
  &F1& 1.0& 0.97&0.97\\ \hline  
 &Precision& 0.69& 0.83&0.71\\ 
 D& Recall& 0.64& 0.71&0.71\\ 
 & F1& \textbf{0.67}& \textbf{0.77}&0.71\\ \hline  
 &Precision& 1.0& 0.94&1.0\\ 
 N& Recall& 1.0& 1.0&1.0\\ 
 & F1& 1.0& 0.91&1.0\\ \hline  
 &Precision& 0.56& 0.67&0.63\\ 
 P& Recall& 0.64& 0.77&0.36\\ 
 & F1& \textbf{0.60}& \textbf{0.71}&0.45\\ \hline  
 &Precision& 0.85& 0.85&0.67\\ 
 S& Recall& 0.79& 0.79&0.71\\ 
 & F1& 0.81& 0.81&0.69\\ \hline  
 &Precision& 0.8& 0.82&0.67\\ 
 V& Recall& 0.8& 0.93&0.93\\ 
 & F1& \textbf{0.8}& \textbf{0.87}&0.78\\ \hline 
\end{tabular}
\end{center}
\end{table}

\begin{figure}
    \centering
    \includegraphics[scale =0.26]{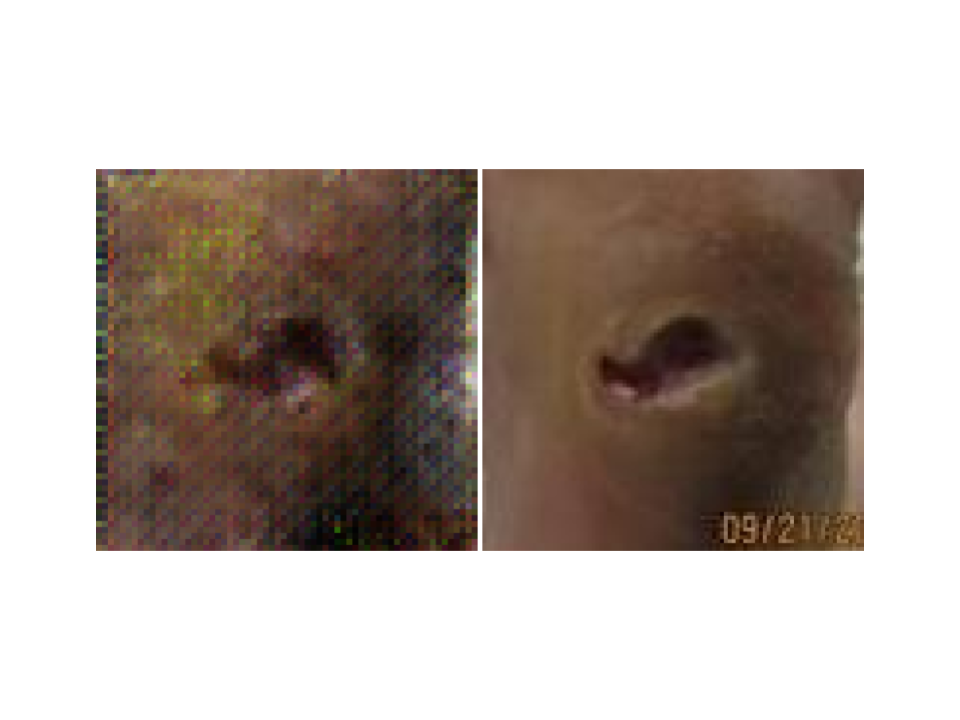}
    \caption{Sample Images Generated by DE-GAN Compared to the Original - \emph{Note. }Left image: generated image of good quality. Right image: real wound image from
original dataset.}
    \label{fig:de-gan-images}
\end{figure}

Due to the high computation (GPU) requirements to train DE-GANs, we first selected one class to experiment. We implemented DE-GAN based augmentation for class D to increase variation in the
training dataset. We did observe good variation in the DE-GAN generated images, as shown in Figure \ref{fig:de-gan-images} for class D. However, we observed that higher quality images did not directly translate into higher F1 score. We are actively working on increasing the number of DE-GAN based augmented images and measure the classification accuracy.

\begin{figure}
    \centering
    \includegraphics[scale=0.25]{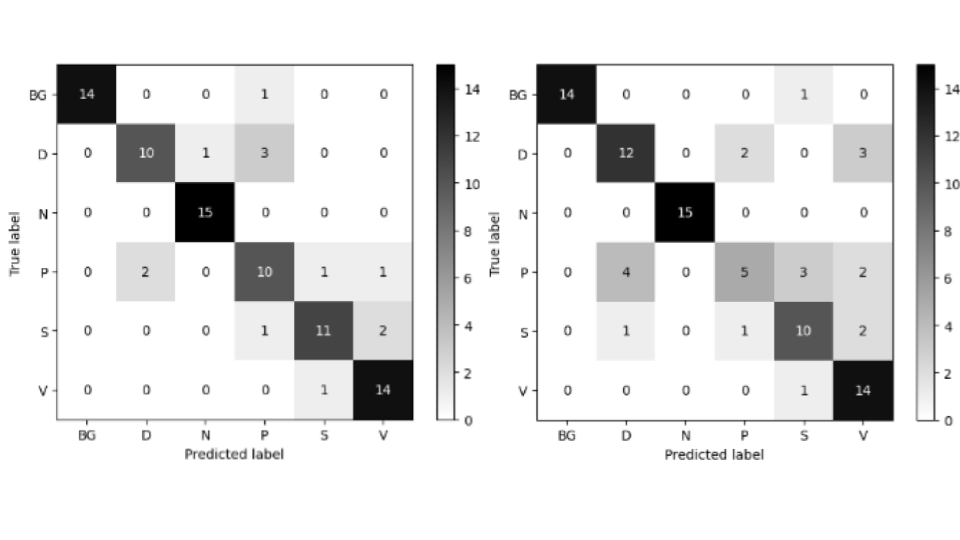}
    \caption{Results with geometric augmentation and GAN Augmentation -
    \emph{Note. }The leftmost image shows the confusion matrix of the
training using geometric augmentation, and the rightmost image shows the
confusion matrix of the training using the DE-GAN augmented dataset.}
    \label{fig:results w/ aug}
\end{figure}

\emph{}

\section{Discussion and Observations}
\label{sec:discussion}

\begin{figure}
    \centering
    \includegraphics[scale=0.125]{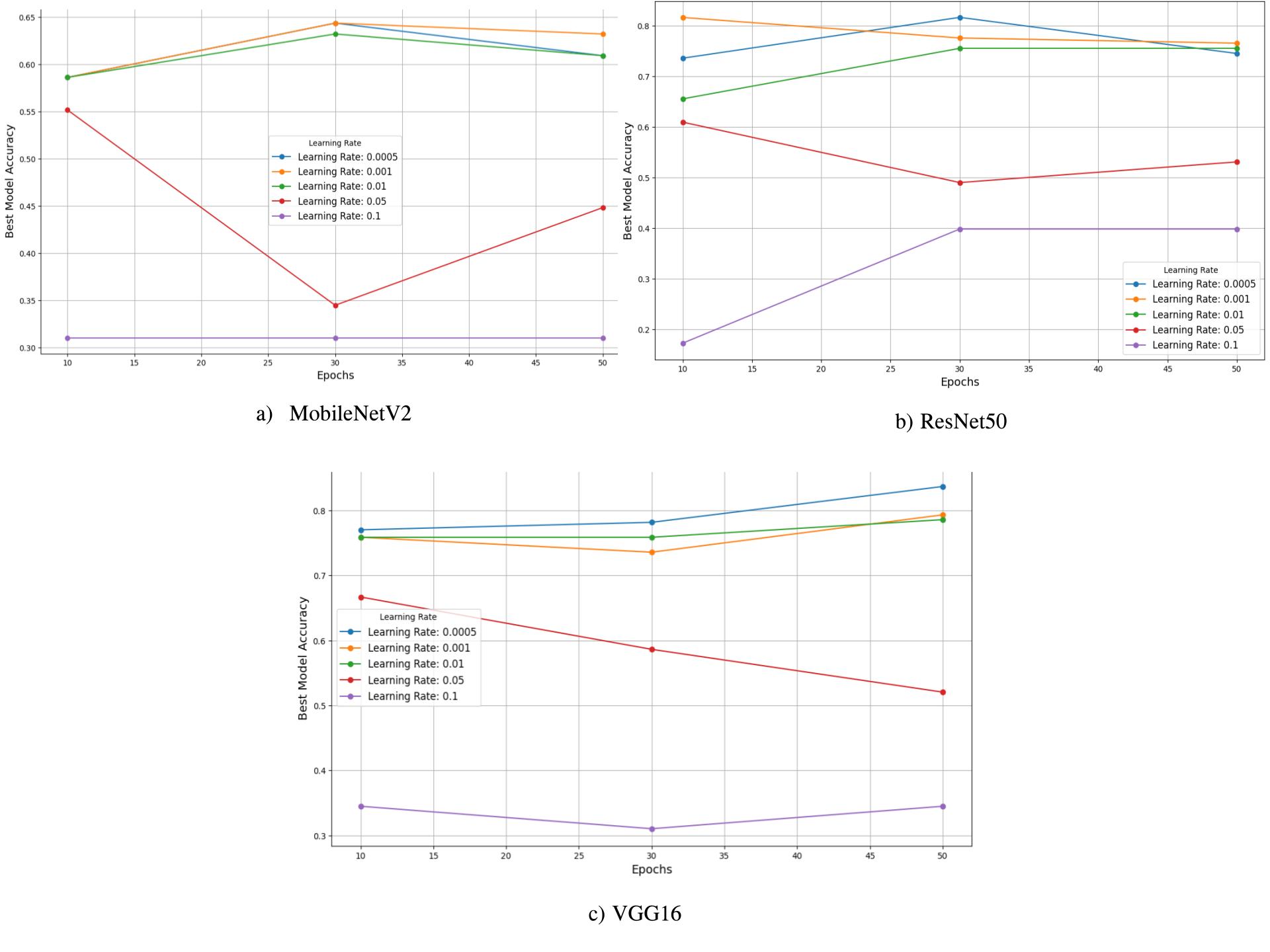}
    \caption{Test Dataset Results with Different Hyperparameter Combinations}
    \label{fig:HPT graphs}
\end{figure}

In this section, we discuss a few key observations from our experiments. Experiments were conducted using three foundational CV models, viz., MobileNetV2, VGG16, ResNet50. Hyper-parameter tuning of these models showed interesting trends in model performance.  When trained for 50 epochs, there was evidence of early stopping during the transfer learning of all three foundation CV models, though these experiments provided the best results. This led us to not train our models past 50 epochs.  All three models were least accurate with 0.01 and 0.05 learning rate, the reason being that the learning rates were much too high and therefore too aggressive to use to train accurate models, as shown in Figure \ref{fig:HPT graphs}. On average, an increase in epochs also induced an increase in accuracy for all three models. 

For the smaller MobileNetV2 model, accuracy remained relatively low across different learning rates when the number of epochs was a low number such as 10 (see Figure \ref{fig:HPT graphs}).  Increasing the number of epochs to 30 showed a general improvement in model accuracy, with the best accuracy observed being 0.64 for the learning rate 0.0005. 

We experimented with a few different variations of the DE-GAN based approach. We explored using colored  vs. black and white wound images, and colored images performed better. We also explored adding more layers to the discriminator of the DE-GAN to increase the depth so that it can provide better signal to the generator, but this did not lead to better quality images.  Finally, we converged on the architecture explained in section \ref{subsec:gan-aug}.

We observed that DE-GANs were very sensitive to large variations in the patterns in the input images and this affected the overall quality of generated images. This learning and the large computational needs of DE-GAN training, led us to focus our efforts on one class (Class D) at a time. Another learning is that DE-GAN requires extensive tuning. Even after focusing on the Class D images, it took extensive hyperparameter search to create realistic wound images. There could be two reasons for this. The first is that the original dataset used to train the GANs had many different patterns (wound on toe versus wound on skin, etc.), which could have inhibited the generator's learning of the dataset. Mode collapse was also very commonly observed during training and led to generation of images that were all very similar to each other.

\section{Conclusion and Future Work}

Our study aimed at investigating data augmentation techniques that can be used for building real-world ML-based wound care systems. To ground the study on a realistic setting, we used  state-of-the-art  CV foundation models as a baseline. We have shown that foundational CV models such as MobileNetV2, ResNet50, and VGG16 can be successfully adapted for wound classification via transfer learning. Building on this, we have shown that geometric data augmentation techniques can provide significant (up to 11\%) improvement on key classes of wounds (D, P and V).  We showed the viability of using DE-GAN based techniques to generate wound images with richer variations as compared to applying only geometric augmentations. Further study is needed to quantify the classification improvements of DE-GANs.  

We are actively exploring the use of DE-GANs to generate a wider variety of wound images and quantifying the classification performance.  Another line of future work is experimentation with other types of GAN and training methods, such as the Wasserstein GAN with gradient penalty \cite{gulrajani2017improved}, which has more stability during training and helps prevent mode collapse.  Layering GAN-based augmentation on top of geometric augmentation is very promising avenue for higher classification performance, and we are actively investigating this. 

\bibliographystyle{ieeetr}
\bibliography{reference}

\end{document}